\documentclass[conference]{IEEEtran}
\usepackage[T1]{fontenc}

\usepackage{calrsfs}
\usepackage{amsmath, amsthm, accents, amssymb, amsfonts, bm,
  mathtools, cases}
\usepackage[cal = boondox,%
  calscaled = 1.1,%
  bb = ams,%
  bbscaled = .94,%
  frakscaled = .97,%
  scr = esstix]{mathalpha}

\usepackage[inline]{enumitem}
\usepackage{graphicx}
\usepackage{hyperref}
\usepackage{booktabs}

\usepackage{caption}
\usepackage{subfig}
\usepackage{textcomp}
\usepackage{etoolbox}
\def\BibTeX{{\rm B\kern-.05em{\sc i\kern-.025em b}\kern-.08em
    T\kern-.1667em\lower.7ex\hbox{E}\kern-.125emX}}

\usepackage[linesnumbered,vlined,ruled,commentsnumbered]{algorithm2e}

\SetAlFnt{\footnotesize}           
\SetAlCapFnt{\footnotesize}        
\SetAlCapNameFnt{\footnotesize}    
\SetAlCapHSkip{0pt}                

\usepackage{cleveref}
\crefname{AlgoLine}{line}{lines}
\Crefname{AlgoLine}{Line}{Lines}
\crefname{algorithm}{Algorithm}{Algorithms}  
\Crefname{algorithm}{Algorithm}{Algorithms}  

\usepackage{url}

\usepackage{siunitx}
\usepackage[textsize=footnotesize]{todonotes}

\usepackage{pgfplots}
\pgfplotsset{compat=newest}
\usepgflibrary[plotmarks]
\usepgfplotslibrary{fillbetween}
\usetikzlibrary{angles, arrows, arrows.meta, shapes, tikzmark, patterns}

\newtheorem{proposition}{Proposition}

\newcommand\given{{\mathbin{}\mid\mathbin{}}}
\newcommand{\IntegerP}{\mathbb{N}}
\newcommand{\IntegerPP}{\mathbb{N}_*}
\newcommand{\Real}{\mathbb{R}}

\newcommand{\PD}{\mathbb{S}_{++}}
\newcommand\SetSymbol[1][]{
  \nonscript\,#1\vert \allowbreak \nonscript\,\mathopen{}}
\DeclarePairedDelimiterX\Set[1]{\lbrace}{\rbrace}%
{ \renewcommand\given{\SetSymbol[\delimsize]} #1 }
\DeclarePairedDelimiterX\norm[1]\lVert\rVert{\ifblank{#1}{\:\cdot\:}{#1}}
\DeclarePairedDelimiterX\innerp[2]{\langle}{\rangle}{#1
  \mathop{}\delimsize\vert\mathop{} #2}

\newcommand\vect[1]{\mathbf{#1}}
\newcommand\vectgr[1]{\bm{#1}}

\newcommand{\Banach}{\mathcal{B}}
\newcommand{\Bellman}{\mathbfcal{T}}

\DeclareMathOperator{\Fix}{Fix}

\DeclareMathOperator{\grad}{grad}

\crefname{theorem}{Theorem}{Theorems}
\crefname{lemma}{Lemma}{Lemmas}
\crefname{thmlisti}{Theorem}{Theorems}
\crefname{figure}{Figure}{Figures}
\crefname{section}{Section}{Sections}
\crefname{proposition}{Proposition}{Propositions}

\newlist{thmlist}{enumerate}{1}
\setlist[thmlist]{label=\textbf{(\roman{*})}, ref=\thetheorem(\roman{*}), noitemsep}

\hypersetup{hidelinks}
\allowdisplaybreaks

\usepackage[%
  backend = biber,%
  maxbibnames = 10,%
  minbibnames = 2,%
  style = ieee,%
  citestyle = numeric-comp,%
  bibencoding = utf8,%
  datamodel = software,%
  defernumbers = true,%
  sortcites=true,%
  doi=false,%
  isbn=false,%
  url=false,%
  eprint=true%
]{biblatex}

\usepackage{tikz}
\usepackage{xcolor}
\usepackage{pgfplots}
\pgfplotsset{compat=newest}
\usepgflibrary{fillbetween}
\usetikzlibrary{angles, arrows, arrows.meta, shapes,
tikzmark, patterns, shapes.geometric}

\pgfdeclareplotmark{halftriangle}{
  \newdimen\dividerwidth
  \dividerwidth=0.6\pgflinewidth
  \newdimen\halfdivider
  \halfdivider=.2\dividerwidth
  \begin{pgfscope}
    \pgfpathmoveto{\pgfpoint{0pt}{\pgfplotmarksize}}
    \pgfpathlineto{\pgfpoint{-\pgfplotmarksize}{-\pgfplotmarksize}}
    \pgfpathlineto{\pgfpoint{\pgfplotmarksize}{-\pgfplotmarksize}}
    \pgfpathclose
    \pgfusepath{clip}
    \pgfpathmoveto{\pgfpoint{-\halfdivider}{\pgfplotmarksize}}
    \pgfpathlineto{\pgfpoint{-\pgfplotmarksize}{-\pgfplotmarksize}}
    \pgfpathlineto{\pgfpoint{-\halfdivider}{-\pgfplotmarksize}}
    \pgfpathclose
    \pgfusepath{fill}
    \pgfsetfillcolor{white}
    \pgfpathmoveto{\pgfpoint{\halfdivider}{\pgfplotmarksize}}
    \pgfpathlineto{\pgfpoint{\pgfplotmarksize}{-\pgfplotmarksize}}
    \pgfpathlineto{\pgfpoint{\halfdivider}{-\pgfplotmarksize}}
    \pgfpathclose
    \pgfusepath{fill}
  \end{pgfscope}
  \pgfpathmoveto{\pgfpoint{0pt}{\pgfplotmarksize}}
  \pgfpathlineto{\pgfpoint{-\pgfplotmarksize}{-\pgfplotmarksize}}
  \pgfpathlineto{\pgfpoint{\pgfplotmarksize}{-\pgfplotmarksize}}
  \pgfpathclose
  \pgfusepath{stroke}
  \begin{pgfscope}
    \pgfsetlinewidth{\dividerwidth}
    \pgfpathmoveto{\pgfpoint{0pt}{\pgfplotmarksize}}
    \pgfpathlineto{\pgfpoint{0pt}{-\pgfplotmarksize}}
    \pgfusepath{stroke}
  \end{pgfscope}
}

\pgfdeclareplotmark{quartedcircle}{
  \begin{pgfscope}
    \pgfpathcircle{\pgfpointorigin}{\pgfplotmarksize}
    \pgfusepath{clip}
    \pgfpathcircle{\pgfpointorigin}{\pgfplotmarksize}
    \pgfusepath{fill}
    \pgfsetfillcolor{white}
    \pgfpathmoveto{\pgfpointorigin}
    \pgfpathlineto{\pgfpoint{\pgfplotmarksize}{0pt}}
    \pgfpathlineto{\pgfpoint{\pgfplotmarksize}{\pgfplotmarksize}}
    \pgfpathlineto{\pgfpoint{0pt}{\pgfplotmarksize}}
    \pgfpathclose
    \pgfusepath{fill}
    \pgfpathmoveto{\pgfpointorigin}
    \pgfpathlineto{\pgfpoint{-\pgfplotmarksize}{0pt}}
    \pgfpathlineto{\pgfpoint{-\pgfplotmarksize}{-\pgfplotmarksize}}
    \pgfpathlineto{\pgfpoint{0pt}{-\pgfplotmarksize}}
    \pgfpathclose
    \pgfusepath{fill}
  \end{pgfscope}
  \pgfpathcircle{\pgfpointorigin}{\pgfplotmarksize}
  \pgfusepath{stroke}
  \pgfpathmoveto{\pgfpoint{-\pgfplotmarksize}{0pt}}
  \pgfpathlineto{\pgfpoint{\pgfplotmarksize}{0pt}}
  \pgfpathmoveto{\pgfpoint{0pt}{-\pgfplotmarksize}}
  \pgfpathlineto{\pgfpoint{0pt}{\pgfplotmarksize}}
  \pgfusepath{stroke}
}

\colorlet{dqn}{green!50!black}
\newcommand{\dqn}{
    \tikz[baseline=-0.5ex]{
        \draw[line width=1pt, color=dqn] (-0.25, 0) -- (0.25, 0);
        \draw[color=dqn] plot[mark size=3pt, mark=square*,
        mark options={line width=1pt}]
        coordinates {(0,0)};
    }
}

\colorlet{ppo}{violet!80!black}
\newcommand{\ppo}{
    \tikz[baseline=-0.5ex]{
        \draw[line width=1pt, color=ppo] (-0.25, 0) -- (0.25, 0);
        \draw[color=ppo] plot[mark size=3pt, mark=triangle*,
        mark options={line width=1pt}]
        coordinates {(0,0)};
    }
}

\colorlet{proposed}{blue!80!black}

\colorlet{klspi}{orange!80!black}

\colorlet{obr}{magenta!80!black}

\colorlet{emgmm}{red!80!black}

\newcommand{\quicksymbol}[2]{
    \tikz[baseline=-0.5ex]{
        \draw[line width=1pt, color=#1] (-0.25, 0) -- (0.25, 0);
        \draw[color=#1] plot[#2]
        coordinates {(0,0)};
    }
}

\newcommand{\dotsymbol}[2]{
    \tikz[baseline=-0.5ex]{
        \draw[color=#1] plot[#2] coordinates {(0,0)};
    }
}

\newcommand{\quickline}[1]{
    \tikz[baseline=-0.5ex]{
        \draw[#1] (-0.25, 0) -- (0.25, 0);
    }
}

\usepackage{amsmath, amssymb, bm}
\usepackage{pgfplots}
\pgfplotsset{compat=1.18}
\usepgfplotslibrary{fillbetween}
\usepackage{siunitx}
\usepackage{pgfplotstable}
\usetikzlibrary{plotmarks}

\usepackage{xcolor}
\definecolor{islamic}{HTML}{009000}
\definecolor{richblack}{HTML}{004040}
\definecolor{pine}{HTML}{01796F}
\definecolor{spring}{HTML}{00FF7F}
\definecolor{celadon}{HTML}{AFE1AF}
\definecolor{olivegreen}{HTML}{808000}
\definecolor{mint}{HTML}{3EB489}

\definecolor{darkorchid}{HTML}{9932CC}
\definecolor{thistle}{HTML}{D8BFD8}
\definecolor{orchid}{HTML}{DA70D6}
\definecolor{lilac}{HTML}{C8A2C8}
\definecolor{ube}{HTML}{8878C3}
\definecolor{darklavender}{HTML}{734F96}

\begin{document}

\title{Online Reinforcement Learning via\\ Sparse Gaussian Mixture Model Q-Functions \vspace{-10pt} }
%
\author{\IEEEauthorblockN{Minh Vu and Konstantinos Slavakis}
\IEEEauthorblockA{Institute of Science Tokyo,
Department of Information and Communications
}
\texttt{vu.d.a5c3@m.isct.ac.jp, slavakis@ict.eng.isct.ac.jp}
}

%
\maketitle

\begin{abstract}
  This paper introduces a structured and interpretable online policy-iteration framework for reinforcement learning
  (RL), built around the novel class of sparse Gaussian mixture model Q-functions (S-GMM-QFs). Extending earlier work
  that trained GMM-QFs offline, the proposed framework develops an online scheme that leverages streaming data to
  encourage exploration. Model complexity is regulated through sparsification by Hadamard overparametrization, which
  mitigates overfitting while preserving expressiveness. The parameter space of S-GMM-QFs is naturally endowed with a
  Riemannian manifold structure, allowing for principled parameter updates via online gradient descent on a smooth
  objective. Numerical experiments show that S-GMM-QFs match
    or even outperform dense deep RL (DeepRL) methods on
    standard benchmarks while using significantly fewer
    parameters. Moreover, they maintain strong performance
    in low-parameter regimes where sparsified DeepRL methods
    fail to generalize.
\end{abstract}


\section{Introduction} \label{sec:intro}

Reinforcement learning (RL) is a machine-learning framework in which an agent controls a system through interactions
with its environment, with the objective of learning an optimal policy that maximizes the expected cumulative reward
resulting from its actions ~\cite{Bertsekas:RLandOC:19, Sutton:IntroRL:18}. RL provides a principled framework for
addressing challenging sequential decision-making problems across diverse domains, including data mining, and the
training of large language models.

A key concept in RL is the \textit{Q-function,} which estimates the expected cumulative reward after the agent takes an
action in a given state under a specific policy. Classical approaches like Q-learning~\cite{watkins92Qlearning} and
SARSA~\cite{singh00sarsa} use tabular representations of Q-functions, computing values for all possible state-action
pairs. While effective for discrete-space problems, these methods become impractical for large or continuous
state-action spaces. To address this issue, RL algorithms based on functional approximation of Q-functions---called
\textit{approximate}\/ RL---have gained significant research interest~\cite{Bertsekas:RLandOC:19}.

In approximate RL, kernel-based methods~\cite{ormoneit02kernel, ormoneit:autom:02, bae:mlsp:11} model
Q-functions in Banach spaces of bounded functions, whereas temporal difference (TD)~\cite{sutton88td}, least-squares
(LSTD)~\cite{xu07klspi, lagoudakis03lspi, regularizedpi:16}, Bellman residual (BR)~\cite{onlineBRloss:16}, and more
recent nonparametric approaches~\cite{vu23rl, akiyama24proximal, akiyama24nonparametric} represent them in reproducing
kernel Hilbert spaces (RKHSs)~\cite{aronszajn50kernels, scholkopf2002learning}. Notwithstanding, nonparametric (kernel)
models typically expand with the amount of data, limiting
scalability in online settings. Sparsification
techniques~\cite{xu07klspi, vu23rl} can mitigate this issue,
but often at the cost of degraded accuracy. Gaussian mixture models (GMMs) have also been used in \textit{distributional}\/ RL, though
their typical role as probability density estimators---either for the joint $p(Q, \vect{s}, a)$~\cite{agostini10gmmrl,
  agostini17gmmrl, sato99em, mannor05rl} or for the conditional $p(Q \given \vect{s}, a)$~\cite{choi19distRL}
distribution---where the Q-function $Q$, state $\vect{s}$, and action $a$ are treated as random variables, 
typically
under strong statistical assumptions such as joint Gaussianity of the sampled data~\cite{mannor05rl}.

A major advance in approximate RL is using deep neural networks as functional Q-function approximators, exemplified by
deep Q-networks (DQNs)~\cite{mnih13dqn, duelingddqn}. While DQNs require many learnable parameters, their parametric
nature provides strong representation power and avoid the model-growth issues of nonparametric methods. Typically, DQNs
learn from past experience~\cite{lin93experience}, enabling exploration beyond the current policy. However, these
usually large ``black-box'' networks are vulnerable to sudden changes in dynamic environments and provide little insight
into the features influencing agent decisions. To reduce model size, sparsification techniques are used: pruning
gradually removes trivial connections from a dense network~\cite{sparseDRL, dense2sparse}, while sparse training fixes a
sparsity pattern from the start and dynamically adjusts
connections~\cite{mocanu17scalable, rigl}. However, even for
these sparsified or pruned models, interpretability remains
limited.

Searching for more expressive power in approximate RL,
\cite{Vu:eusipco:25, minh25tsp} introduced Gaussian mixture model
Q-functions (GMM-QFs) to represent Q-functions as weighted
sums of multivariate Gaussian components with learnable
weights, mean vectors and covariance matrices.
Unlike distributional RL that uses GMMs to model
probability density functions, GMM-QFs are used directly as a parametric
functional approximator of Q-functions.
The number of mixture components is user-specified, allowing control over model
complexity and mitigating the curse of dimensionality in nonparametric approaches.
In~\cite{Vu:eusipco:25, minh25tsp}, GMM-QFs are employed
within an offline, on-policy policy-iteration (PI) framework via
BR, exploiting Riemannian
geometry of the parameter space. 

Building upon this foundation,
the present work proposes an
\textit{online} and \textit{off-policy} PI framework that
learns from streaming data
while simultaneously constructing an experience buffer. This
buffer is actively exploited to enhance both exploration and
mitigate biased nature of on-policy approaches. 
Furthermore, to improve the scalability and interpretability of
GMM-QFs~\cite{Vu:eusipco:25, minh25tsp}, a novel class of
sparse (S-)GMM-QFs is introduced via Hadamard
overparametrization. This formulation enables explicit and transparent
sparsification, yielding a compact mixture of a few
geometrically meaningful Gaussians learned through online
gradient-descent methods on a smooth objective defined over the
Riemannian manifold of model parameters.
This approach contrasts
with DeepRL approaches, where sparsification typically targets network structures
and often lacks interpretability. 
Numerical tests on benchmark datasets demonstrate that
S-GMM-QFs maintain strong performance while substantially
reducing model complexity. Results are comparable
to---and in some cases surpass---those of DeepRL models
which use significantly more parameters, while effectiveness
is retained even in low-parameter-count regimes where sparse DeepRL
methods fail to generalize. Due to space limitations, detailed
derivations, convergence analyses, and additional numerical tests will be reported in a separate publication.

\section{Sparse GMM Q-Functions}\label{sec:model}

\subsection{RL notation and basics on Bellman mappings}

Let $\mathfrak{S} \subset \Real^{D_s}$ denote a \textit{continuous}\/ state space, with state vector $\vect{s} \in
\mathfrak{S}$, for some $D_s \in \IntegerPP$ ($\IntegerPP$ denotes the set of all positive integers). The discrete
action space is denoted by $\mathfrak{A}$, with action $a
\in \mathfrak{A}$. For convenience, cardinality of action
space $N_a \coloneqq \lvert
\mathfrak{A} \rvert < \infty$. An agent at state $\vect{s} \in \mathfrak{S}$ takes action $a \in \mathfrak{A}$ and
transits to a new state $\vect{s}'\in \mathfrak{S}$ under an unknown transition probability $p(\vect{s}' \given
\vect{s}, a)$ with a reward $r(\vect{s}, a)$. The Q-function $Q(\cdot, \cdot) \colon \mathfrak{S} \times \mathfrak{A}
\to \Real \colon (\vect{s}, a) \mapsto Q(\vect{s}, a)$ stands for the long-term cumulative reward achievable if the
agent selects action $a$ in state $\vect{s}$. Following~\cite{Bertsekas:RLandOC:19}, a (deterministic) policy
$\mu(\cdot)$ maps a state to an action, as in $\mu(\cdot)\colon \mathfrak{S} \to \mathfrak{A} \colon \vect{s} \mapsto
\mu(\vect{s})$. Denote also the set of all mappings from
$\mathfrak{S}$ to $\mathfrak{A}$ by $\mathcal{M}$. Let
also $\overline{1,N} \coloneqq \{ 1, \ldots, N\}$.

The Q-function is determined by the Bellman
mapping~\cite{Bertsekas:RLandOC:19} through relationship
between immediate rewards and the discounted future values.
More precisely, when Q-functions are drawn from the functional space
$\Banach$---typically the Banach space of essentially bounded functions~\cite{Bertsekas:RLandOC:19}---the (classical)
Bellman mapping $\Bellman^{\diamond}_{\mu} \colon \Banach \to \Banach \colon Q \mapsto \Bellman^{\diamond}_{\mu} Q$ for
a policy $\mu(\cdot)$ is defined as: $\forall (\vect{s}, a)$,
\begin{align}
  (\Bellman_{\mu}^{\diamond} Q)(\vect{s}, a)
  & \coloneqq r( \vect{s}, a ) + \alpha \mathbb{E}_{\vect{s}^{\prime} \given
    (\vect{s}, a)} \bigl[ Q(\vect{s}^{\prime}, \mu(\vect{s}^{\prime}))
    \bigr]\,, \label{Bellman.standard.mu}
\end{align}
where $\mathbb{E}_{\vect{s}' \given (\vect{s}, a)} [\cdot]$ is the conditional expectation operator with respect to the
next state $\vect{s}'$ conditioned on $(\vect{s}, a)$, and $\alpha \in [0, 1)$ being the discount factor. A greedy
  version of \eqref{Bellman.standard.mu} is the Bellman mapping $\Bellman^{\diamond} \colon \Banach \to \Banach \colon Q
  \mapsto (\Bellman^{\diamond} Q)(\vect{s}, a) \coloneqq r( \vect{s}, a ) + \alpha \mathbb{E}_{\vect{s}^{\prime} \given
    (\vect{s}, a)} [ \max_{a^{\prime}\in \mathfrak{A}} Q(\vect{s}^{\prime}, a^{\prime}) ]$.

The fixed-point set of $\Bellman^{\diamond}_{\mu}$ is defined as $\Fix \Bellman^{\diamond}_{\mu} \coloneqq \Set{Q \in
  \Banach \given Q = \Bellman^{\diamond}_{\mu} Q}$. It is well-known that identifying a fixed point $Q^{\diamond}_{\mu}
\in \Fix \Bellman^{\diamond}_{\mu}$ plays a central role in computing optimal policies that maximize cumulative
rewards. Usually, $\alpha \in [0, 1)$ to assure that \eqref{Bellman.standard.mu} becomes a strict contraction, hence
  $\Fix\Bellman^{\diamond}_{\mu}$ is a singleton~\cite{Bertsekas:RLandOC:19, hb.plc.book}. It follows from
  \eqref{Bellman.standard.mu} that identifying $Q^{\diamond}_{\mu}$ requires access to $\mathbb{E}_{\vect{s}' \given
    (\vect{s}, a)} [\cdot]$, which, however, is typically unavailable to RL agents in practice.

\subsection{Sparse GMM-QFs}

Extending~\cite{minh25tsp}, the proposed functional class of S-GMM-QFs is defined as follows: for user-defined $K, J
\in \IntegerPP$,
\begin{align}
  \mathcal{Q}_K \coloneqq \Bigl \{
  Q & \colon \mathfrak{S} \times \mathfrak{A} \to \Real \notag\\[-15pt]
  & \colon (\vect{s}, a) \mapsto Q(\vect{s}, a) \coloneqq \sum_{k=1}^{K} \prod_{j=1}^{J}
  \upsilon_{k,j}(a) \mathscr{G}(\vect{s} \given \vect{m}_k, \vect{C}_k) \notag\\[-5pt]
  & \Big \vert\, \upsilon_{k,j}(a) \in \Real, \vect{m}_k \in \Real^{D_s}, \vect{C}_k \in \PD^{D_s}
  \Bigl \} \,, \label{eq:GMMQF}
\end{align}
where $\mathscr{G}(\vect{s} \given \vect{m}_k, \vect{C}_k) \coloneqq \mathscr{G}_k (\vect{s}) \coloneqq \exp[ -(\vect{s}
  - \vect{m}_k)^{\intercal} \vect{C}_k^{-1} (\vect{s} - \vect{m}_k)]$, and $\PD^{D_s}$ denotes the set of all $D_s
\times D_s$ positive definite matrices, while $\intercal$ stands for vector/matrix transposition.

To justify the design, the functional class \eqref{eq:GMMQF} enjoys the following ``universal-approximation'' property.

\begin{proposition}\label{prop:dense}
  The union $\cup_{K=1}^{\infty} \mathcal{Q}_K$ is dense in the space of all square-(Lebesgue)-integrable functions on
  $\mathfrak{S} \times \mathfrak{A}$.
\end{proposition}

The proposed S-GMM-QFs extend the model of~\cite{minh25tsp}, where a single scalar weight $\xi_{k}(a)$ was used in place
of $\prod_{j=1}^{J} \upsilon_{k,j}(a)$. Define the $K \times N_a$ matrix $\vectgr{\Upsilon}_j$ (recall $N_a = \lvert
\mathfrak{A} \rvert$) such that $[\vectgr{\Upsilon}_j]_{k,a} \coloneqq \upsilon_{k,j}(a)$. Then, the $K \times N_a$
matrix $\vectgr{\Xi}$, whose entries are $[\vectgr{\Xi}]_{k,a} \coloneqq \xi_{k}(a) \coloneqq \prod_{j=1}^J
\upsilon_{k,j}(a)$, satisfies the Hadamard factorization $\vectgr{\Xi} = \odot_{j=1}^J \vectgr{\Upsilon}_j$, where
$\odot$ denotes the Hadamard product. This overparametrization of $\vectgr{\Xi}$ by $\Set{ \vectgr{\Upsilon}_j }_j$
promotes sparsity, thereby enhancing efficiency and robustness~\cite{hoff2017lasso, li2023tail,
  ziyin2023spred, hadamard_sparse}. More specifically, it has been shown that such overparametrization, combined with
the smooth regularizer $\sum_{j=1}^J \norm{ \vectgr{\Upsilon}_j }_{\text{F}}^2$ (with $\norm{\cdot}_{\text{F}}$ being
the Frobenius norm), induces the nonconvex quasi-norm $\norm{\vectgr{\Xi}}_{2/J}^{2/J}$ for $J>2$, which yields sparser
solutions for $\vectgr{\Xi}$ than classical $\ell_1$-norm regularization~\cite{hadamard_sparse}. This Hadamard
overparametrization offers a transparent and interpretable mechanism for sparsifying the Gaussian mixture in
\eqref{eq:GMMQF}, as it enables users to readily identify the Gaussian components that contribute most to the Q-function
estimate. Finally, overparametrization, through increasing the number of degrees of freedom, may further improve
approximation quality~\cite{hoff2017lasso, li2023tail, ziyin2023spred, hadamard_sparse}.

Let now the $D_s \times K$ matrix $\vect{M} \coloneqq [\vect{m}_1, \dots, \vect{m}_K]$. Then, the learnable parameters
can be collected as $\vectgr{\Omega} \coloneqq (\vectgr{\Upsilon}_1, \dots, \vectgr{\Upsilon}_J, \vect{M}, \vect{C}_1,
\dots, \vect{C}_K)$, so that each $\vectgr{\Omega}$
specifies a single S-GMM-QF in \eqref{eq:GMMQF}.
Altogether, these parameters define the parameter space $\mathfrak{M}_K \coloneqq
(\Real^{K \times N_a})^J \times \Real^{D_s \times K} \times (\PD^{D_s})^{K}$, which becomes a Riemannian manifold
because it is the Cartesian product of Riemannian manifolds~\cite{RobbinSalamon:22,
  Absil:OptimManifolds:08}. Consequently, the tangent space to $\mathfrak{M}_K$ at $\vectgr{\Omega}$ becomes
$T_{\vectgr{\Omega}} \mathfrak{M}_K = (\Real^{K \times N_a})^J \times \Real^{D_s \times K} \times T_{\vect{C}_1}
\PD^{D_s} \times \ldots \times T_{\vect{C}_K} \PD^{D_s}$, where $T_{\vect{C}_k} \PD^{D_s}$ stands for the tangent space
to $\PD^{D_s}$ at $\vect{C}_k$, known to be the set of all $D_s \times D_s$ symmetric matrices~\cite{RobbinSalamon:22,
  Absil:OptimManifolds:08}.

Motivated by the significance of the fixed point $Q^{\diamond}_{\mu} \in \Fix \Bellman^{\diamond}_{\mu}$, the widely
used Bellman-residual (BR) approach~\cite{onlineBRloss:16, regularizedpi:16} estimates $Q^{\diamond}_{\mu}$ by any
minimizer in $\arg \min_{\overline{Q}} (1/T) \sum_{t=1}^{T} [ \overline{Q}(\vect{s}_t, a_t) - r_t - \alpha
  \overline{Q}(\vect{s}'_t, \mu(\vect{s}'_t)) ]^2$ of an empirical loss, defined via the sample dataset $\mathcal{D}
\coloneqq \Set{(\vect{s}_t, a_t, r_t \coloneqq r(\vect{s}_t, a_t), \vect{s}'_t)}_{t=1}^{T}$. This empirical-loss
formulation compensates for the typical inaccessibility of $\mathbb{E}_{\vect{s}' \given (\vect{s}, a)} [\cdot]$ in
\eqref{Bellman.standard.mu}, a situation encountered in nearly all practical scenarios. Inspired by the BR approach, as
well as by the classical temporal-difference (TD) strategy~\cite{sutton88td}, this study builds upon the following
smooth empirical loss, parameterized by the sample dataset $\mathcal{D}$ and a user-defined Q-function $\overline{Q}$:
$\forall \vectgr{\Omega} \in \mathfrak{M}_K$,%
\begin{subequations}\label{BRM.TD}
\begin{align}
  L_{\mu} (\vectgr{\Omega}; \overline{Q}, \mathcal{D})
  \coloneqq \frac{1}{T} \sum_{t=1}^{T} \Big[
    & \sum_{k=1}^{K} \prod_{j=1}^{J} \upsilon_{k,j} (a_t) \mathscr{G}_k(\vect{s}_t) \notag\\
    & - r_t - \alpha 
    \overline{Q} (\vect{s}'_t, \mu (\vect{s}'_t)) \Big]^2 \,. \label{eq:BRM.param}
\end{align}
Note that only the column vectors of $\Set{ \vectgr{\Upsilon}_j }_{j=1}^J$ corresponding to $\Set{a_t}_{t=1}^{T}$ enter
\eqref{eq:BRM.param} through $\mathcal{D}$. The remaining columns, which do not appear in $\mathcal{D}$, make no
contribution to the loss.
Building on the earlier discussion on promoting sparsity
via Hadamard overparametrization,
the empirical loss is further extended to incorporate a smooth regularization term, leading to the following objective:
$\forall \vectgr{\Omega}\in \mathfrak{M}_K$,
\begin{align}
  \mathscr{L}_{\mu} (\vectgr{\Omega}; \overline{Q}, \mathcal{D})
  \coloneqq L_{\mu}(\vectgr{\Omega}; \overline{Q}, \mathcal{D}) + \underbrace{ \rho \sum\nolimits_{j=1}^{J} \norm{
    \vectgr{\Upsilon}_j }_{\text{F}}^2 }_{ \mathscr{R}(\vectgr{\Omega}) } \label{eq:regBRM}\,,
\end{align}%
\end{subequations}%
where $\rho > 0$ is a user-defined regularization coefficient. It is worth emphasizing that $\mathscr{L}_{\mu}$ is
rendered smooth via Hadamard overparametrization, in sharp contrast to the non-smooth losses typically encountered in
sparsity-inducing $\ell_p$-norm regularized minimization tasks with $0 < p \leq 1$.

\section{Online Approximate PI via S-GMM-QF\MakeLowercase{s}}\label{sec:online-PI}

This study focuses on the online-learning setting, in which streaming data are presented sequentially to the RL agent at
each time instance $n \in \IntegerPP$. The agent also makes decisions sequentially and updates its policies through a
policy-iteration (PI) scheme whose algorithmic index of steps is aligned with the time index $n$.

More precisely, at each time instance $n$, the controlled system provides its current state (data) $\vect{s}_n$ to the
agent, which then selects an action $a_n \coloneqq \mu_n(\vect{s}_n)$ according to the current policy $\mu_n$. The
environment returns a reward (feedback) $r_n$ in response to this action, and the system transitions to the next state
$\vect{s}_n^{\prime}$, which becomes $\vect{s}_{n+1} \coloneqq \vect{s}_n^{\prime}$ for the subsequent time
instance. This interaction is summarized by the tuple $(\vect{s}_n, a_n, r_n, \vect{s}_{n+1})$. The agent accumulates
such ``experience'' in a buffer $\mathcal{B}_n$, of user-defined capacity $B \in \IntegerPP$, which stores all
previously observed tuples and is updated according to $\mathcal{B}_{n+1} = \mathcal{B}_n \cup \Set{(\vect{s}_n, a_n,
  r_n, \vect{s}_{n+1})}$. When the buffer exceeds its capacity, i.e., $\lvert \mathcal{B}_{n+1} \rvert > B$, its oldest
tuple is discarded.

The novel S-GMM-QFs are integrated into a classical online approximate policy-iteration (PI) scheme---see
\cref{algo:gmmq}. Like any standard PI procedure, \cref{algo:gmmq} consists of two steps: policy evaluation and policy
improvement. In the policy-evaluation step, the current policy $\mu_n$ guides the update of the S-GMM-QF estimate using
data from the experience buffer $\mathcal{B}_n$. In the policy-improvement step, the agent updates its policy based on
the updated S-GMM-QF estimate. To underscore the online nature of the scheme, the formulations in \cref{sec:model} are
indexed by the discrete time step $n$.

At each time step $n$, given the current S-GMM-QF estimate
$Q_n$, a dataset $\mathcal{D}_n \coloneqq \Set{
    (\vect{s}_t^{(n)}, a_t^{(n)}, r_t^{(n)},
    \vect{s}'_t{}^{(n)} ) }_{t=1}^{T}$ is uniformly sampled from the
experience buffer $\mathcal{B}_n$ to construct the loss
$\mathscr{L}_{\mu} ( \cdot ; Q_n, \mathcal{D}_n)$ according
to \eqref{BRM.TD}. Since this loss is defined on the
Riemannian manifold $\mathfrak{M}_K$, and the data arrive
via online streaming, an off-the-shelf Riemannian
steepest-gradient-descent approach is employed. In
particular, the Riemannian gradient $\grad \mathscr{L}_{\mu_n} 
= \grad L_{{\mu}_n} +
\grad \mathscr{R}$~\cite{Absil:OptimManifolds:08} is
computed, and the parameters are updated using Riemannian
(R-)Adam~\cite{RAdam}, due to its significant performance
over plain-vanilla gradient descent.
Given $\vectgr{\Pi}_{n-1}
\in T_{\vectgr{\Omega}_{n-1}}\mathfrak{M}_K$ and $\sigma_{n-1} \in \Real$, the update
estimate $\vectgr{\Omega}_{n+1}$ is obtained as
follows~\cite{RAdam}: $\forall n \in \IntegerPP$, 
\begin{subequations}\label{eq:update.Omega}
  \begin{alignat}{2}
    \vectgr{\Pi}_n & {} \coloneqq {} && \beta_1 \varphi_{\vectgr{\Omega}_{n-1} \to \vectgr{\Omega}_{n}}
    (\vectgr{\Pi}_{n-1}) \notag \\
    &&& + (1 - \beta_1) \grad \mathscr{L}_{\mu_n} (\vectgr{\Omega}_n; Q_n, \mathcal{D}_n)
    \,, \label{eq:parallel.trans} \\
    \sigma_{n} & \coloneqq && \beta_2 \sigma_{n-1} \notag \\ 
    &&& + (1 - \beta_2)\norm{\grad \mathscr{L}_{\mu_n}
      (\vectgr{\Omega}_n; Q_n, \mathcal{D}_n)}^2_{\vectgr{\Omega}_n} \,, \\
    \vectgr{\Omega}_{n+1} & \coloneqq && R_{\vectgr{\Omega}_n} [ - \gamma \vectgr{\Pi}_{n} (1 - \beta_2^n) /
      ( \sigma_{n} (1 - \beta_1^n) ) ] \label{eq:update.5} \,,
  \end{alignat}
\end{subequations}
with user-defined exponential decay rates $\beta_1, \beta_2 \in (0, 1)$,
learning rate $\gamma$.
The update involves \textit{parallel
transport}~\cite{RobbinSalamon:22, Absil:OptimManifolds:08}
$\varphi_{\vectgr{\Omega}_{n-1} \to \vectgr{\Omega}_{n}}
\colon T_{\vectgr{\Omega}_{n-1}}\mathfrak{M}_K \to
T_{\vectgr{\Omega}_{n}}\mathfrak{M}_K$
in~\eqref{eq:parallel.trans}, which transfers
tangent vectors between successive tangent spaces to
preserve consistency of $\vectgr{\Pi}_n$ on
$T_{\vectgr{\Omega}_n}\mathfrak{M}_K$, and a
\textit{retraction mapping}~\cite{RobbinSalamon:22,
Absil:OptimManifolds:08} $R_{\vectgr{\Omega}_n}:
T_{\vectgr{\Omega}}\mathfrak{M}_K \to
\mathfrak{M}_K$
in~\eqref{eq:update.5}, which projects tangent-space
updates back onto $\mathfrak{M}_K$. Due to
space constraints, detailed formulations and derivations of
Riemannian optimization are deferred to future
work.

\begin{algorithm}[t!]
\caption{Online approximate PI via S-GMM-QFs}\label{algo:gmmq}
\SetAlgoLined

Arbitrarily initialize 
    $\vectgr{\Omega}_0 \in \mathfrak{M}_K$, thus
    by~\eqref{eq:GMMQF} $Q_0
    \in \mathcal{Q}_K$, 
    policy $\mu_0 \in \mathcal{M}$, and capacity $B \in \IntegerPP$ of experience buffer\;
$n\gets 0$, $\vectgr{\Pi}_0 \gets \vect{0}$, $\sigma_0 \gets 0$, experience buffer $\mathcal{B}_0 \gets \emptyset$\;
Environment starts at an initial state $\vect{s}_0 \in \mathfrak{S}$\;

\While{$n \in \IntegerP$ \label{line:while.loop}}{%
    Tuple $(\vect{s}_n, a_n, r_n, \vect{s}_{n+1})$ becomes available to the agent\;
    $\mathcal{B}_{n+1} \gets \mathcal{B}_n \cup \{(\vect{s}_n, a_n, r_n, \vect{s}_{n+1})\}$\;
    \lIf{$\lvert \mathcal{B}_{n+1} \rvert > B$}{discard oldest tuple}

    Sample dataset $\mathcal{D}_n$ from $\mathcal{B}_{n+1}$\;
    Define $\mathscr{L}_{\mu_n}(\cdot; Q_n, \mathcal{D}_n)$ by~\eqref{BRM.TD}\;

    \textbf{Policy evaluation:}

    \label{line:policy.eval} Compute $\grad \mathscr{L}_{\mu_n}(\vectgr{\Omega}_n; Q_n, \mathcal{D}_n)$\label{line:compute.gradient}\;
    Update $\vectgr{\Omega}_{n+1}$ by \eqref{eq:update.Omega}\;
    Define $Q_{n+1}\in\mathcal{Q}_K$ by
    $\vectgr{\Omega}_{n+1}$ via~\eqref{eq:GMMQF};

    \textbf{Policy improvement:} $\mu_{n+1}(\vect{s}) \gets \arg\max_{a \in \mathfrak{A}} Q_{n+1}(\vect{s}, a)$\;

    \lIf{task terminated}{reset $\vect{s}_{n+1} \gets \vect{s}_0$}

    $n \gets n+1$ and go to line~\ref{line:while.loop}\;
}

\end{algorithm}

\section{Numerical tests} \label{sec:tests}

\begin{figure}[t]
    \centering
    \subfloat[Acrobot]{\includegraphics[width=.49\columnwidth]{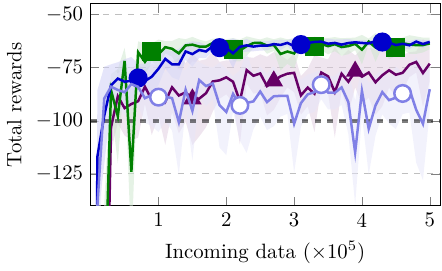}
    \label{subfig:acrobot-before}
    }
    \subfloat[Lunar
    lander]{\includegraphics[width=.49\columnwidth]{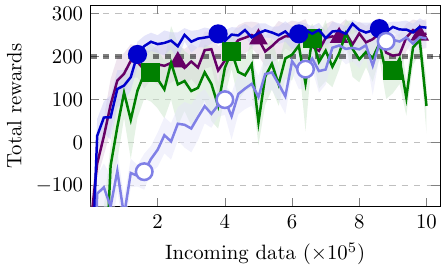}
    \label{subfig:lunarlander-before}
    }
    \caption[]{Performance of dense models. Curve markers:~\cref{algo:gmmq}
    ($K=50$):~\quicksymbol{proposed!50}{mark=*, mark
    size=3pt, mark options={line width=1pt, fill=white}}, 
    ($K=500$):~\quicksymbol{proposed}{mark=*, mark
    size=3pt, mark options={line width=1pt}},
    dense DQN~\cite{mnih13dqn}:~\dqn,
    dense PPO~\cite{PPO}:\ppo. \quickline{dashed,
    line width=2pt, opacity=0.5}: desired total rewards.
    } \label{fig:before}
\end{figure}

\begin{figure}[t]
    \centering
    \subfloat[Acrobot]{
        \includegraphics[width=.85\columnwidth]{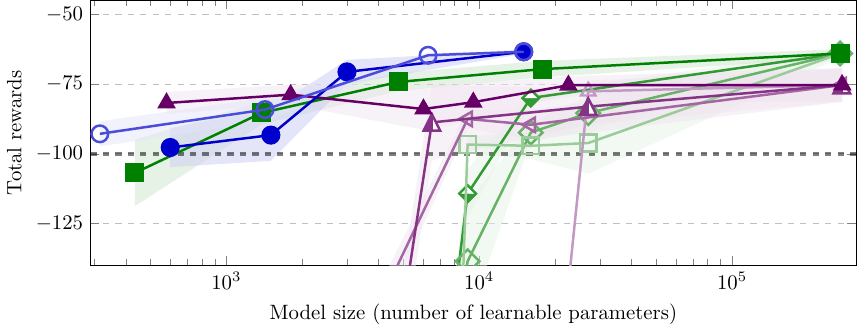}
        \label{subfig:acrobot-after}
    }
    \\
    \subfloat[Lunar lander]{
        \includegraphics[width=.85\columnwidth]{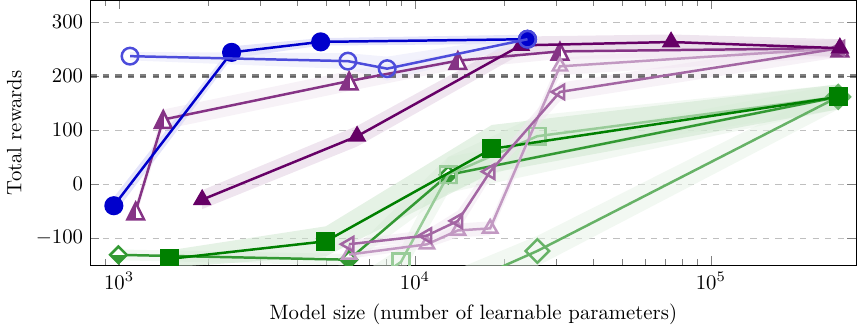}
        \label{subfig:lunarlander-after}
    }
    \caption[]{
        Performance against number of parameters.
        Curve markers:~\cref{algo:gmmq} (dense,
        $K$-tuning):~\quicksymbol{proposed}{mark=*, mark
        size=3pt},~\cref{algo:gmmq} (sparse from $K=500$,
        Hadamard-overparametrized):~\quicksymbol{proposed}{mark=o,
        mark size=3pt, mark options={line width=1pt}};
        dense DQNs~\cite{mnih13dqn} (model-tuning):\dqn,
        static
        DQN~\cite{sparseDRL}:\quicksymbol{dqn!40}{mark=square,
        mark options={line width=1pt}, mark size=3pt},
        Prune-DQN~\cite{dense2sparse}:\quicksymbol{dqn!80}{mark=halfsquare*,
        mark options={line width=1pt}, mark size=3pt},
        SET-DQN~\cite{mocanu17scalable}:\quicksymbol{dqn!60}{mark=square,
        mark options={line width=1pt, rotate=45}, mark
        size=3pt};
        dense PPO~\cite{PPO} (model-tuning):\ppo, static
        PPO~\cite{sparseDRL}:\quicksymbol{ppo!40}{mark=triangle,
        mark options={line width=1pt}, mark size=3pt},
        Prune-PPO~\cite{dense2sparse}:\quicksymbol{ppo!80}{mark=halftriangle,
        mark options={line width=1pt}, mark size=3pt},
        SET-PPO~\cite{mocanu17scalable}:\quicksymbol{ppo!60}{mark=triangle,
        mark options={line width=1pt, rotate=90}, mark
        size=3pt}. \quickline{dashed,
        line width=2pt, opacity=0.5}: desired total rewards.
    }
    \label{fig:perform-vs-sparse}
\end{figure}

Two RL benchmarks with finite-cardinality action spaces---the acrobot and the
lunar lander~\cite{gym}---are selected to validate the
proposed framework against the popular online DeepRL
methods, including DQN~\cite{mnih13dqn} and PPO~\cite{PPO}, as well as against their sparsified variants:
\begin{enumerate*}[label=\textbf{(\roman*)}]
\item pruned (dense-to-sparse)~\cite{dense2sparse},
\item static sparse~\cite{sparseDRL}, training a fixed sparse network, and
\item SET~\cite{mocanu17scalable}, which trains and updates
    sparsity via cosine decay criterion.
\end{enumerate*}
PPO~\cite{PPO} employs two separate neural networks for the policy
(actor)
and Q-function (critic),
whereas DQN, similarly to \cref{algo:gmmq}, models only the
Q-function.
Performance is measured via the cumulative reward (vertical
axis) achieved by the learned policy $\mu_n$ against number
of streaming data $n$. To illustrate the effect of
sparsification, performance of trained models is also plotted
vs. model size (number of parameters).
Algorithms are evaluated every \num{1e4} incoming data for
acrobot and every \num{2e4} for lunar lander, 
with results averaged over \num{20} independent episodes
using a separate test emulator. 
Average performance over multiple training sessions is
recorded in~\cref{fig:before,fig:perform-vs-sparse}.
Numerical
comparisons of the proposed framework's predecessor with kernel-based and distributional
RL methods are reported in~\cite{minh25tsp}.

The ``acrobot'' is an underactuated two-link robot, where
the objective is to swing the free end from the
downward to the upright position in as few steps as
possible. The state is \num{4}-dimensional, including joint angles and angular velocities of
both links, and the action
space has \num{3} options: accelerate left, right, or do
nothing. The agent receives a reward
of $0$ only upon reaching the upright configuration and $-1$ otherwise. Episodes start from the downward position and
finish upon reaching the goal or after \num{500} steps. 
The
``lunar lander'' is a spacecraft control task aiming
to land a lunar module on a designated pad. The
state space has \num{8} dimensions, capturing the lander's positions, velocity,
orientation, angular velocity and contact of two legs. The action space has \num{4}
discrete options: do nothing, fire the left, main or right
engine. The environment-defined reward encourages safe,
upright landings and penalizes crashes, drifts, and
excessive fuel usage. 
Episodes start with the
lander descending from above and terminate upon a successful
landing, a crash, leaving the allowed area or after
\num{1000} steps.


Hyperparameters for~\cref{algo:gmmq} are chosen as follows: number of Hadamard weight components $J=3$, and maximum
capacity $B$ of $\mathcal{B}_n$ is \num{1e4} for acrobot,
and \num{1e5} for lunar lander, with the size of $\mathcal{D}_n$ set to $T = 64$. 
To retain importance of updated information in objective function when sampling from experience data, the latest
incoming data tuple is always included in $\mathcal{D}_n$.  The learning rate of R-Adam~\cite{RAdam} is set to
\num{1e-4}. 
Following the suggestion of \cite{sparseDRL}, sparsification
is applied 
on a dense network with 2 hidden
layers of size \num{512} to represent Q-functions and PPO
critic, whereas the actor of PPO is modeled separately
using a dense network of 2 \num{64}-neuron hidden layers.
Performance of DeepRL methods under several model configurations,
achieving by tuning hidden-layer sizes,
is also reported in~\cref{fig:perform-vs-sparse}.
Adam~\cite{Adam} is used to train the neural networks, with
the learning rate \num{1e-3} for DQNs and \num{3e-4} for
PPOs, as well as for all of their sparse training variants.
Note that, contrary to sparse DeepRL methods, S-GMM-QFs
control sparsity via the regularizer $\rho$, as shown
in~\cref{fig:perform-vs-sparse} for $\rho=\Set{0, 0.005,
0.01, 0.05}$ recorded, with $\rho=0$ corresponding to the
dense model ($K=500$).


\Cref{fig:before} reports the performance of all methods
under dense configurations.
In~\cref{subfig:acrobot-before},~\cref{algo:gmmq} ($K=500$)
converges rapidly and exhibits stable performance.
PPO~\cite{PPO} shows fast initial improvement but settles on
suboptimals policies, while DQN~\cite{mnih13dqn} achieves
competitive performance. In the more challenging task of lunar
lander (\cref{subfig:lunarlander-before}), S-GMM-QFs
($K=500$) outperform DeepRL baselines with faster and more
stable learning, whereas DQN~\cite{mnih13dqn} fails to score
the desired behavior. Note that, S-GMM-QFs use far fewer
parameters than DeepRL models, and as shown
in~\cref{fig:perform-vs-sparse}, retain their performance
under high sparsification in both control tasks. 
In contrast, dense DQN and PPO with ``tiny''
networks achieve the desired
performance on acrobot (\cref{subfig:acrobot-after}), but
degrade significantly in low-parameter regimes on lunar
lander (\cref{subfig:lunarlander-after}).
Interestingly, both
tasks favor S-GMM-QFs with
sparsity over their dense counterparts at comparable
parameter budgets, highlighting the effectiveness of structured sparsification
via Hadamard overparametrization; while DeepRL methods
suffer sharp performance drop as sparsified to match
parameter count of S-GMM-QFs.


\begin{figure}[t]
    \centering
    \subfloat[S-GMM-QFs, without sparsification (\num{500}
    Gaussians)]{
        \includegraphics[width=.98\columnwidth]{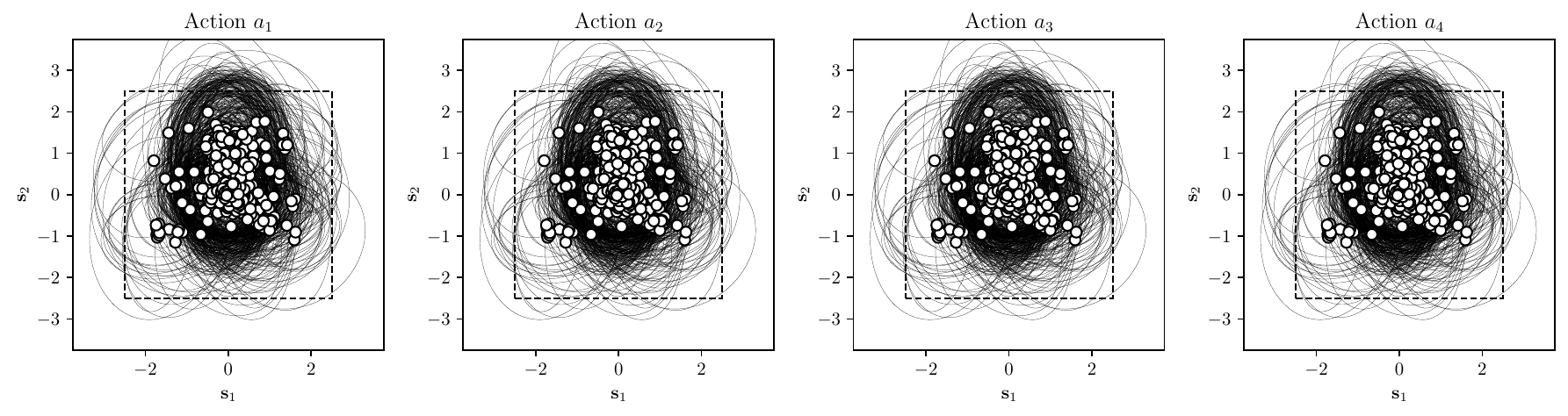}
    }
    \\
    \subfloat[S-GMM-QFs (18 Gaussians, sparsified from
    $\num{500}$ Gaussians)]{
        \includegraphics[width=.98\columnwidth]{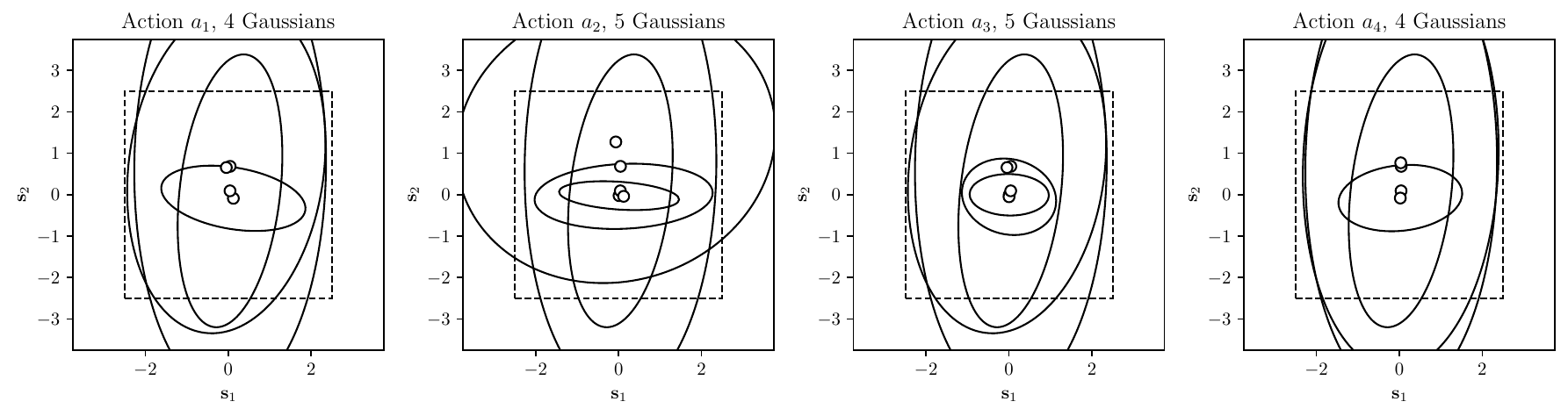}
    }
    \caption[]{Effect of sparsification via Hadamard
    overparametrization ($J=3\,,\rho=0.05$), lunar lander dataset. Multivariate GMMs are
    projected onto lander's positions. Dashed box denotes
    the state domain. \dotsymbol{black}{mark=o, mark
    options={line width=1pt}}: Gaussian
    centers.}
    \label{fig:interpret}
\end{figure}
\Cref{fig:interpret} compares distribution of Gaussians in
dense and sparsified S-GMM-QFs. \Cref{algo:gmmq} removes
redundant components while preserving expressiveness,
yielding models supported by only a few dominant Gaussians.
This sparsity produces a more compact and interpretable
representation, where individual component contributions are
clearer, without compromising overall performance.
\footnotesize
\printbibliography[title = {\normalsize\uppercase{References}}]

\end{document}